\documentclass[runningheads]{llncs}
\usepackage{graphicx}
\usepackage{relsize}
\usepackage{caption}
\usepackage{amsmath}
\usepackage{tikz}
\usepackage{gensymb}
\usepackage{hyperref}
\usepackage[colorinlistoftodos]{todonotes}
\begin{document}
\title{High quality ultrasonic multi-line transmission through deep learning}
\author{Sanketh Vedula\inst{1} \and Ortal Senouf\inst{1} \and Grigoriy Zurakhov\inst{1} \and Alex Bronstein\inst{1}  \and \\ Michael Zibulevsky\inst{1} \and Oleg Michailovich\inst{2} \and  Dan Adam\inst{1} \and Diana Gaitini\inst{3}}
\authorrunning{Vedula et al.}

 \institute{Technion - Israel Institute of Technology \\ \email{\{sanketh,senouf\}@campus.technion.ac.il}\and University of Waterloo, Canada \and Rambam Health Care Campus and Faculty of Medicine, Technion}
%
%
\maketitle              
\begin{abstract}
Frame rate is a crucial consideration in cardiac ultrasound imaging and 3D sonography. Several methods have been proposed in the medical ultrasound literature aiming at accelerating the image acquisition. In this paper, we consider one such method called \textit{multi-line transmission} (MLT), in which several evenly separated focused beams are transmitted simultaneously. While MLT reduces the acquisition time, it comes at the expense of a heavy loss of contrast due to the interactions between the beams (cross-talk artifact). In this paper, we introduce a data-driven method to reduce the artifacts arising in MLT. To this end, we propose to train an end-to-end convolutional neural network consisting of correction layers followed by a constant apodization layer. The network is trained on pairs of raw data obtained through MLT and the corresponding \textit{single-line transmission} (SLT) data. Experimental evaluation demonstrates significant improvement both in the visual image quality and in objective measures such as contrast ratio and contrast-to-noise ratio, while preserving resolution unlike traditional apodization-based methods. We show that the proposed method is able to generalize well across different patients and anatomies on real and phantom data.
\end{abstract}

\section{Introduction}\label{intro}
Medical ultrasound is a wide-spread imaging modality due to its high temporal resolution, lack of harmful radiation and cost-effectiveness, which distinguishes it from other modalities such as MRI and CT. 
High frame rate ultrasound is highly desirable for the functional analysis of rapidly moving organs, such as the heart. For a given angular sector size and acquisition depth, the frame rate is limited by the speed of sound in soft tissues (about $1540$ m/s). The frame rate depends on the number of transmitted beams needed to cover the field of view; thus, it can be increased by lowering the number of the transmitted events. One such method termed \emph{multi-line acquisition} (MLA) or \emph{parallel receive beamforming} (PRB) employs a smaller number of wide beams in the transmission, and constructs a multiple numbers of beams in the reception \cite{PRB1991},\cite{PRB1984}. The drawbacks of the method include block-like artifacts in images, reduced lateral resolution, and reduced contrast \cite{rabinovich2013multi}. Another high frame-rate method, \emph{multi-line transmission} (MLT), employs a simultaneous transmissions of a multiple number of narrow beams focused in different directions \cite{mallart1992improved},\cite{drukarev1993beam}. Recently reinvented, this method suffers from a high energy content due to the simultaneous transmissions \cite{santos2015acoustic}, and from cross-talk artifacts on both the transmit and receive, caused by the interaction between the beams \cite{tong2013multi},\cite{tong2014multi}. 

\begin{figure}[t]
\centering
\includegraphics[width = 0.8\textwidth]{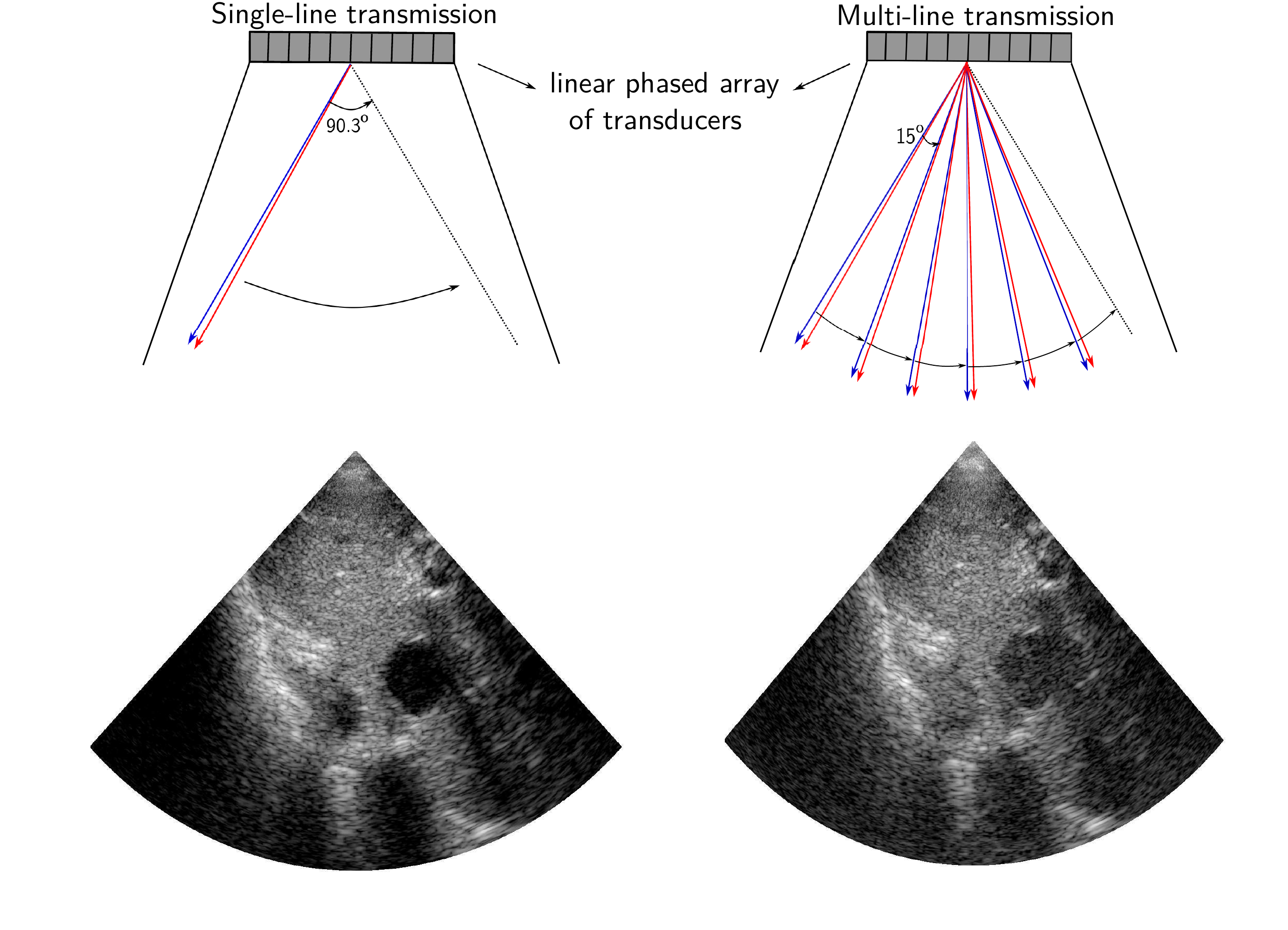}
\label{mltslt}
\vspace{-0.5cm}
\caption{Single- (left) vs. Multi- (right, with MLT factor of $6$) line transmission procedures and their corresponding ultrasound scans. Severe drop in contrast can be observed in the case of MLT. Blue and red lines correspond two consecutive transmissions.}
\vspace{-0.1cm}
\end{figure}

Over the years, numerous methods were proposed to deal with those artifacts, including constant \cite{tong2013multi},\cite{tong2014multi} and adaptive \cite{rabinovich2015multi}, \cite{zurakhov2018multi} apodizations, by allocating different frequency bands to different transmissions \cite{demi2012parallel},\cite{demi2015vitro}, and by using a tissue harmonic mode \cite{prieur2013correspondence}. The filtered delay-multiply-and-sum beamforming (F-DMAS) \cite{matrone2015delay} was proposed in the context of MLT in \cite{matrone2017high}, demonstrating better artifact rejection, higher contrast ratio (CR) and lateral resolution compared to MLT beamformed with delay-and-sum (DAS) and Tukey apodization on receive, at expense of lower contrast-to-noise ratio (CNR). Finally, short-lag F-DMAS for MLT was studied in \cite{matrone2018spatial}, demonstrating a contrast improvement for higher maximum-lag values, and resolution and speckle-signal-to-noise ratio (sSNR) improvements for lower lag values, at the expense of decreased MLT cross-talk artifact rejection.
By using a simulated $2-$MLT, it was demonstrated in \cite{prieur2013correspondence} that the tissue harmonic imaging mode provides images with a lower transmit cross-talk artifact as compared to the fundamental harmonic imaging. However, the receive cross-talk artifact still requires correction. In the present study, we demonstrate that similarly to the fundamental harmonic, the cross-talk is more severe in the tissue harmonic mode for higher MLT configurations, which is manifested by a lower contrast. 

Convolutional neural networks (CNN) were introduced for the processing of ultrasound acquired data in order to generate a high quality plane wave compounding with a reduced number of transmissions \cite{gasse2017high} as well as for fast despeckling, and CT-quality image generation \cite{vedula2017towards} during the post-processing stage. In a parallel effort, \cite{senouf2018high} demonstrated the effectiveness of CNNs in improving MLA quality in ultrasound imaging. To the best of our knowledge, ours is the first attempt to use CNN in MLT ultrasound imaging.\\
\\
\textit{Contributions. } In this work, we propose an end-to-end CNN-based approach for MLT artifact correction. We train a convolutional neural network consisting of an encoder-decoder architecture followed by a constant apodization layer. The network is trained with dynamically focused element-wise data obtained from {\it{in-vivo}} scans in an simulated MLT configuration with the objective to approximate the corresponding single-line transmission (SLT) mode. We demonstrate the performance of our method both qualitatively and quantitatively using metrics such as CR and CNR. Finally, we validate that the trained model generalizes well to different patients, different anatomies, as well as to phantom data.

\section{Methods}
{ \textbf{MLT simulation.  }} Acquisition of the real MLT data is a complicated task that requires a highly flexible ultrasound system. Fortunately, MLT can be faithfully simulated using the data acquired in a single-line transmit (SLT) mode by summation of the received data prior to the beamforming stage, as was done in \cite{prieur2013correspondence} and \cite{rabinovich2015multi} for the fundamental and tissue harmonic modes. It should be noted that while MLT can be simulated almost perfectly in a fundamental harmonic case, there is a restriction in the tissue harmonic mode due to the nonlinearity of its forward model. It was shown in \cite{prieur2013correspondence} that in the tissue harmonic mode, the summation of the data sequentially transmitted in two directions provides a good enough approximation for the simultaneous transmission in the same directions if the MLT separation angle is above $15^{\circ}$. The assumption behind the present study is that this approximation holds for a higher MLT number, as long as the separation angle remains the same, since the beam profile between two beams is mainly affected by those beams. For this reason, $4-$MLT and $6-$MLT with separation angles of 22.6$^{\circ}$ and 15.06$^{\circ}$, respectively, were used in this study.  

\begin{figure}[t]
\vspace{-0.5cm}
\centering
\includegraphics[width=\textwidth,height=0.4\textheight]{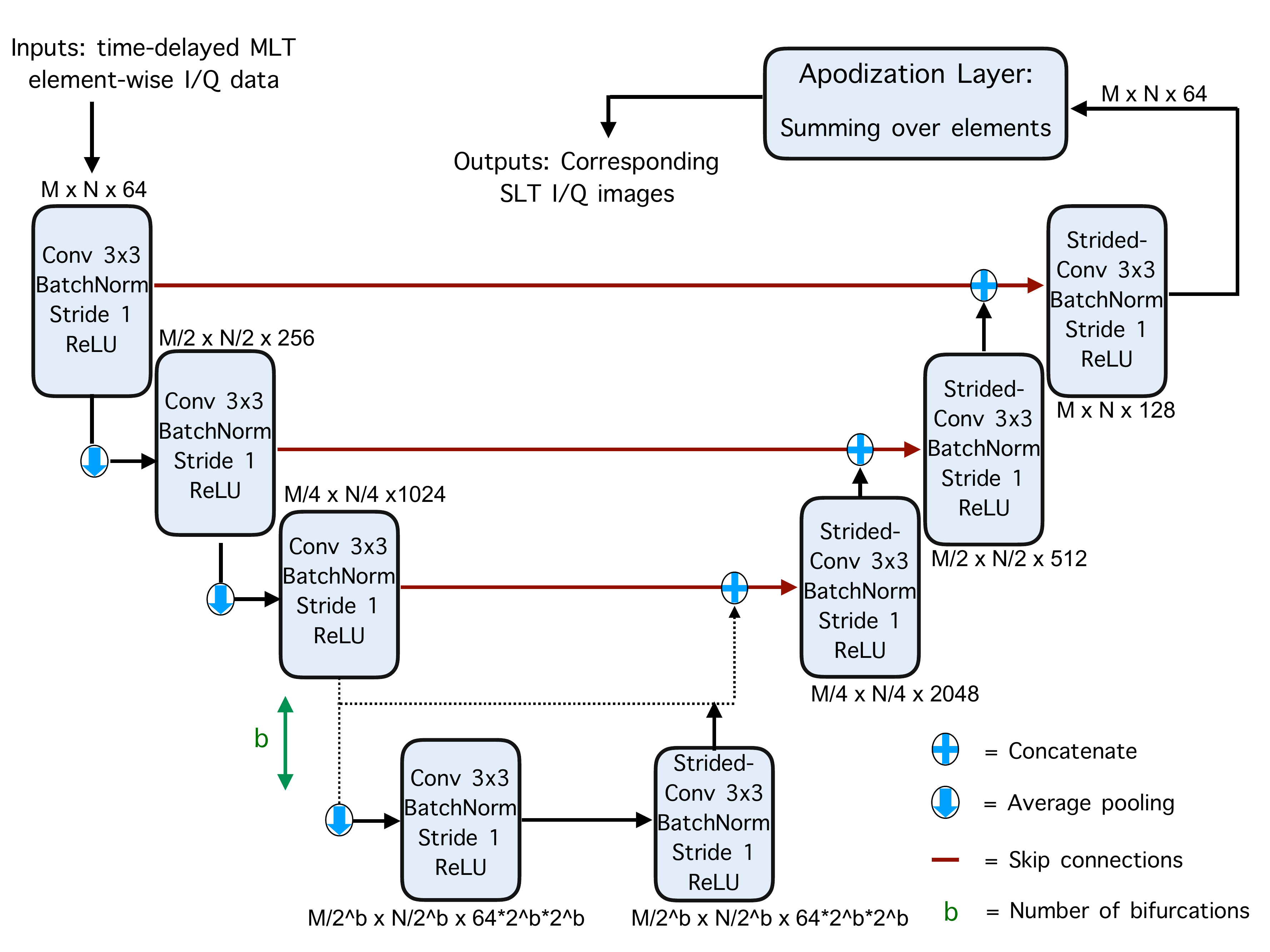}
\label{mlt_pipeline_fig}
\vspace{-0.2mm}
\caption{CNN-based MLT artifact correction pipeline. For all the experiments within this paper: $M=696, N=180, b=5$}
\vspace{-0.2mm}
\end{figure}

Clinical use mandates the use of lower excitation voltage in real MLT, implemented in a standard way \cite{santos2015acoustic}, due to patient safety considerations, which will affect the generation of the tissue harmonic and signal-to-noise ratio (SNR). The latter issue can probably be adressed by the CNNs, that are capable of learning denoising tasks, as has been demonstrated in \cite{zhang2017beyond}. It should be noted, that alternative implementations of MLT were proposed in \cite{santos2015acoustic}, allowing a safer application of the method. However, to the best of our knowledge, no study was performed concerning impact of those methods on image quality.
Nevertheless, this study focuses on testing whether the MLT artifact can be corrected using CNN, while the optimization of the number of simultaneous transmissions in the tissue harmonic mode is beyond its scope. 
\\
\\
\textbf{Data acquisition. }
For the purpose of the study, we chose imaging of quasi-static internal organs, such as bladder, prostate, and various abdominal structures, since the simulated MLT of the rapidly moving organ may alter the cross-talk artifact. The study was performed with the data acquired using a GE ultrasound system, scanning $6$ healthy human volunteers and a tissue mimicking phantom (GAMMEX Ultrasound 403GS LE Grey Scale Precision Phantom). The tissue harmonic mode was chosen for this study, being a common mode for cardiac imaging, with a contrast resolution that is superior to the fundamental harmonic, at either $f_0$ or $2f_0$.
The scans were performed in a transversal plane by moving a probe in a slow longitudinal motion in order to reduce the correlation in the training data acquired from the same patient. The acquisition frame rate was $18$ frames per second.
Excitation sinusoidal pulses of $2.56$ cycles, centered around \textit{f\textsubscript{0}}=$1.6$ MHz, were transmitted using a $64$-element phased array probe with the pitch of $0.3$mm. No apodization was used on transmit. On receive, the tissue harmonic signal was demodulated (I/Q) at 3.44 MHz and filtered. A $90.3^{\circ}$ field-of-view (FOV) was covered with $180$ beams. In the case of MLT, the signals were summed element-wise with the appropriate separation angles. Afterward, both SLT and MLT were dynamically focused and summed. In the simulated MLT mode the data were summed after applying a constant apodization window (Tukey, $\alpha=0.5$) as the best apodization window in \cite{tong2013multi},\cite{tong2014multi}. At training, non-apodized MLT and SLT data were presented to the network as the input and the desired output, respectively.
\label{acquisition}
\\
\\
\textbf{Improving MLT quality using CNNs. }  
As mentioned earlier, traditional methods tackle the cross-talk artifacts by performing a linear or non-linear processing of a time-delayed element-wise data to reconstruct each pixel in the image. In this work, we propose to replace the traditional pipeline of MLT artifact correction with an end-to-end CNN, as depicted in Figure \ref{mlt_pipeline_fig}.  \\
\textit{Network architecture. } The proposed network resembles a fully-convolutional autoencoder (albeit different training regime), consisting of $10$ layers with symmetric skip connections from each layer in the upsampling track to each layer within the downsampling track \cite{mao2016image}. All the convolutions set to $3 \times 3$, stride $1$ and the non-linearities are set to ReLU. Downsampling is performed through average pooling and strided convolutions are used for upsampling. The network accepts time-delayed phase-rotated element-wise I/Q data from the transducer obtained through MLT as the input. \\
\textit{Apodization stage. }  A \textit{constant} apodization layer is introduced following the downsampling and upsampling tracks. It is implemented as $1 \times 1$ convolutions consisting of $64$ channels which are applied element-wise and initialized with a boxcar function (window of ones). The layer can be implement any constant apodization such as Tukey or Hann windows.\\
\textit{Training. }
Following the apodization at the last output stage, the network outputs an artifact-corrected I/Q image. At training, SLT I/Q image are used both to generate a simulated MLT input data as well as the corresponding SLT (artifact-free) reference output. The network is trained as a regressor minimizing the $L_1$ discrepancy between the predicted network outputs and the corresponding ground-truth SLT data. The loss is minimized using Adam optimizer \cite{kingma2014adam}, with the  learning rate set to $10^{-4}$. The training data were acquired as described in previous sections. A total of $750$ frames from the acquired sequences were used for training. The input to the network is a MLT I/Q image of size $696 \times 180 \times 64$ (depth $\times$ lines $\times$
elements) and the output is an SLT-like I/Q image data of size $696 \times 180$ (depth $\times$ lines). The training is performed separately for the I and Q components of the image.

\section{Experimental Evaluation}
\textbf{Settings. }
 In order to evaluate the performance of the networks trained on $4-$ and $6-$MLT setups, we consider a test set consisting of two frames from the bladder and one frame from a different anatomy acquired from a patient excluded from the training set, and a phantom frame.  While all the chosen test frames were unseen during training, the latter two frames portray different image classes that were not part of the training set. The data were acquired as described in Section \ref{acquisition}. Evaluation was conducted both visually and quantitatively using CR and CNR objective measures as defined in \cite{matrone2018spatial}.
\\
\\
\textbf{Results and discussion. } Figure \ref{KidneyFigs} and S$1$-$2$ (in the supplementary material)
depict the SLT groundtruth, and the artifact-corrected $4-$ and $6-$MLT images. Figure \ref{KidneyFigs} demonstrates a number of anatomical structures in abdominal area, as depicted by the arrows. The CNN processing has restored the CR loss caused by the MLT cross-talk artifact for the $4-$MLT, and improved the CR by a $9.8$ dB for the $6-$MLT, as measured for aorta (yellow contour) and a background region (magenta contour).  S$1$ demonstrates structures in a tissue mimicking phantom, such as anechoic cyst (the black circle marked by a yellow rectangle) and number of a point reflectors. Finally, S$2$ demonstrates a bladder (large dark cavity)  and a prostate, located beneath it, scanned in a transversal plane. The output of our CNN was compared to the MLT image with Tukey ($\alpha=0.5$) window apodization on receive, a common method to the attenuation of the receive cross-talk artifact.

\begin{figure}[t]
\begin{minipage}[]{\linewidth}
	\begin{tabular}{ c@{\hskip 0.001\textwidth}c@{\hskip 0.001\textwidth}c@{\hskip 0.001\textwidth}c} 
		\includegraphics[width = 0.33\textwidth]{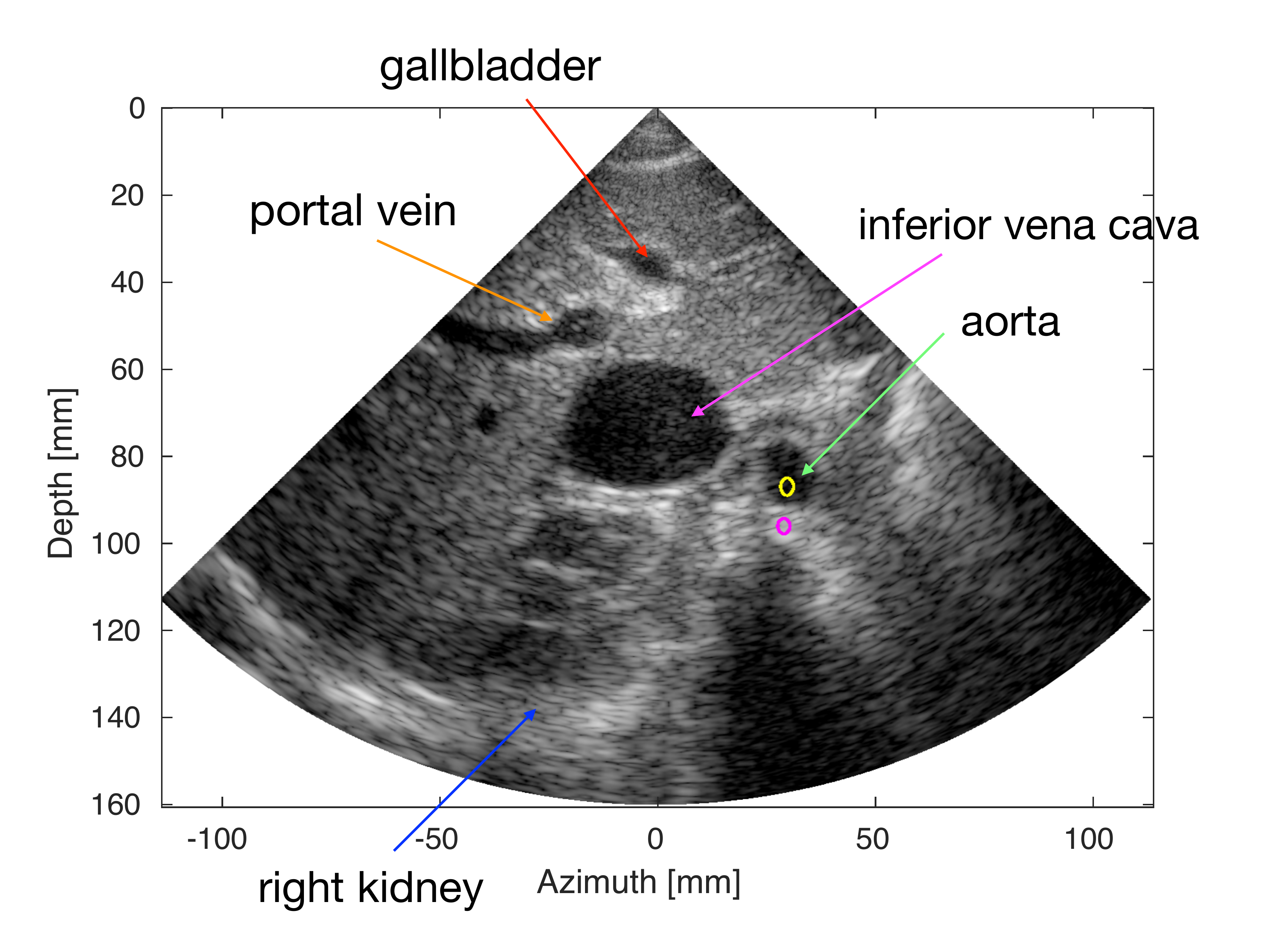} &
		\includegraphics[width = 0.33\textwidth]{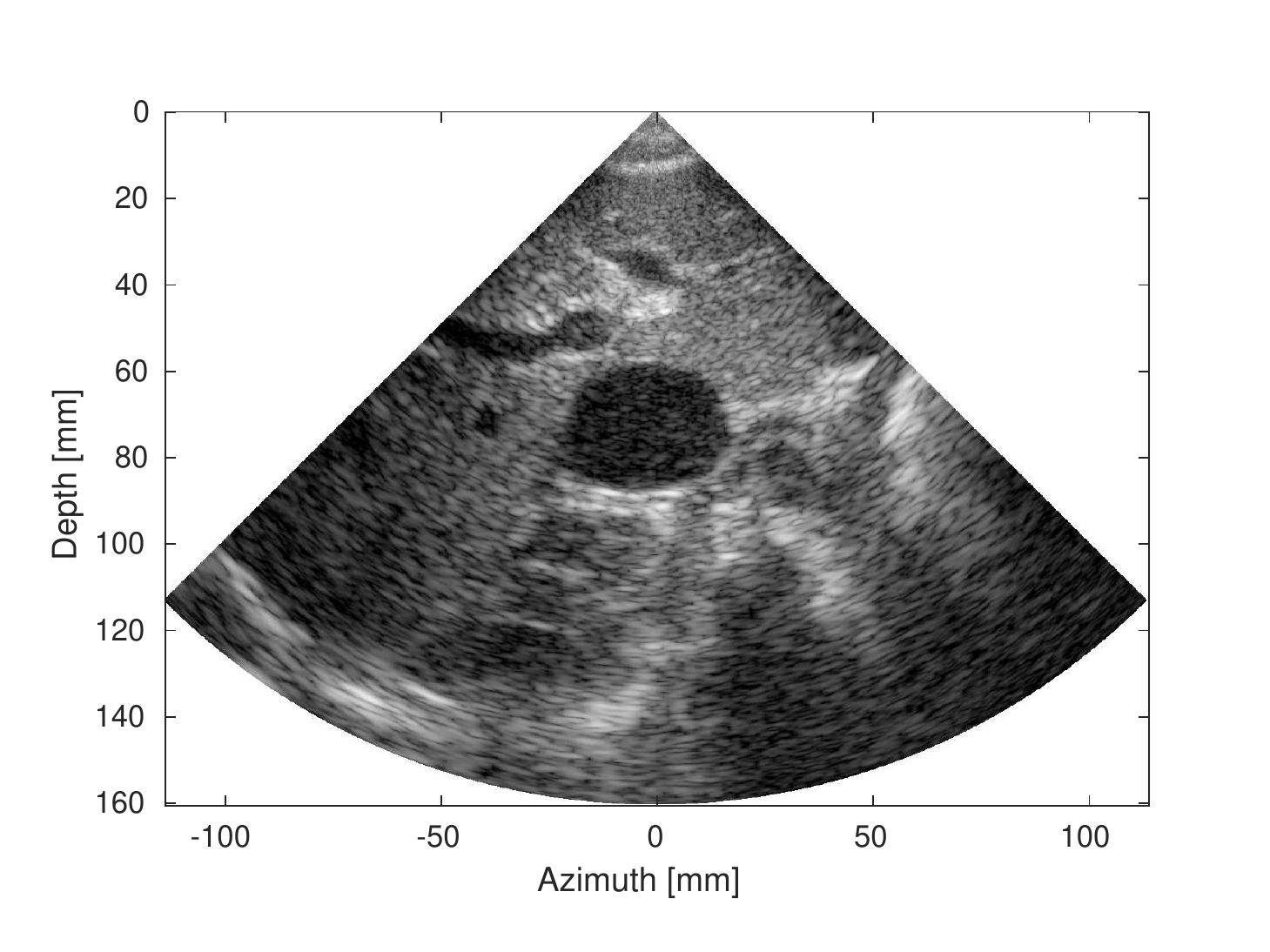} &
		\includegraphics[width = 0.33\textwidth]{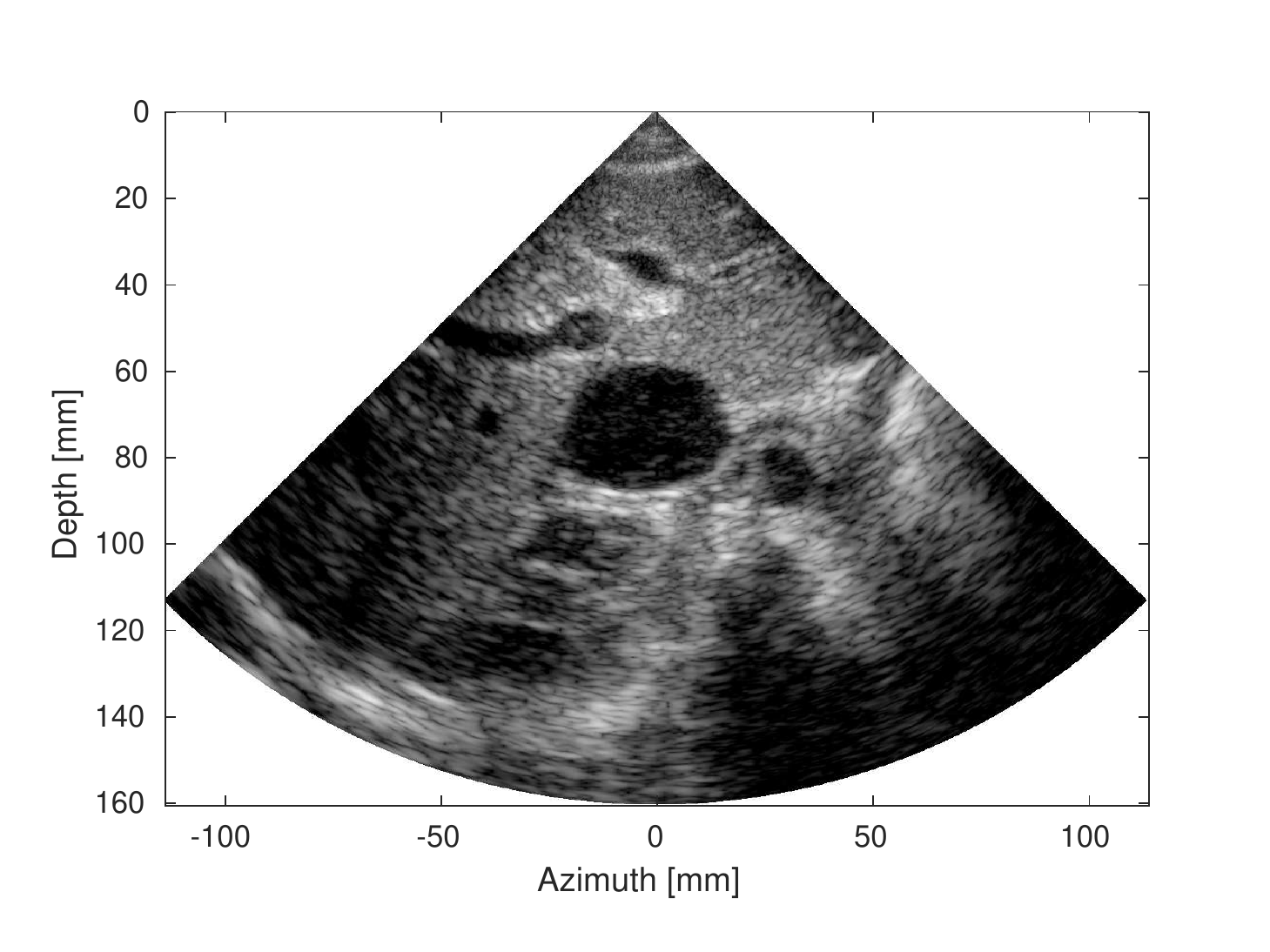} &
	       \\ {\smaller (a) SLT} & {\smaller (b) $4-$MLT, (Tukey, $\alpha$=0.5)} &  {\smaller (c) $4-$MLT, CNN}  \\
               {\smaller CNR=2.52, CR=-37.83dB} & {\smaller CNR=2.57, CR=-27.93dB } & {\smaller CNR=2.59, CR=-37.87dB} &\\ 
           &
		\includegraphics[width = 0.33\textwidth]{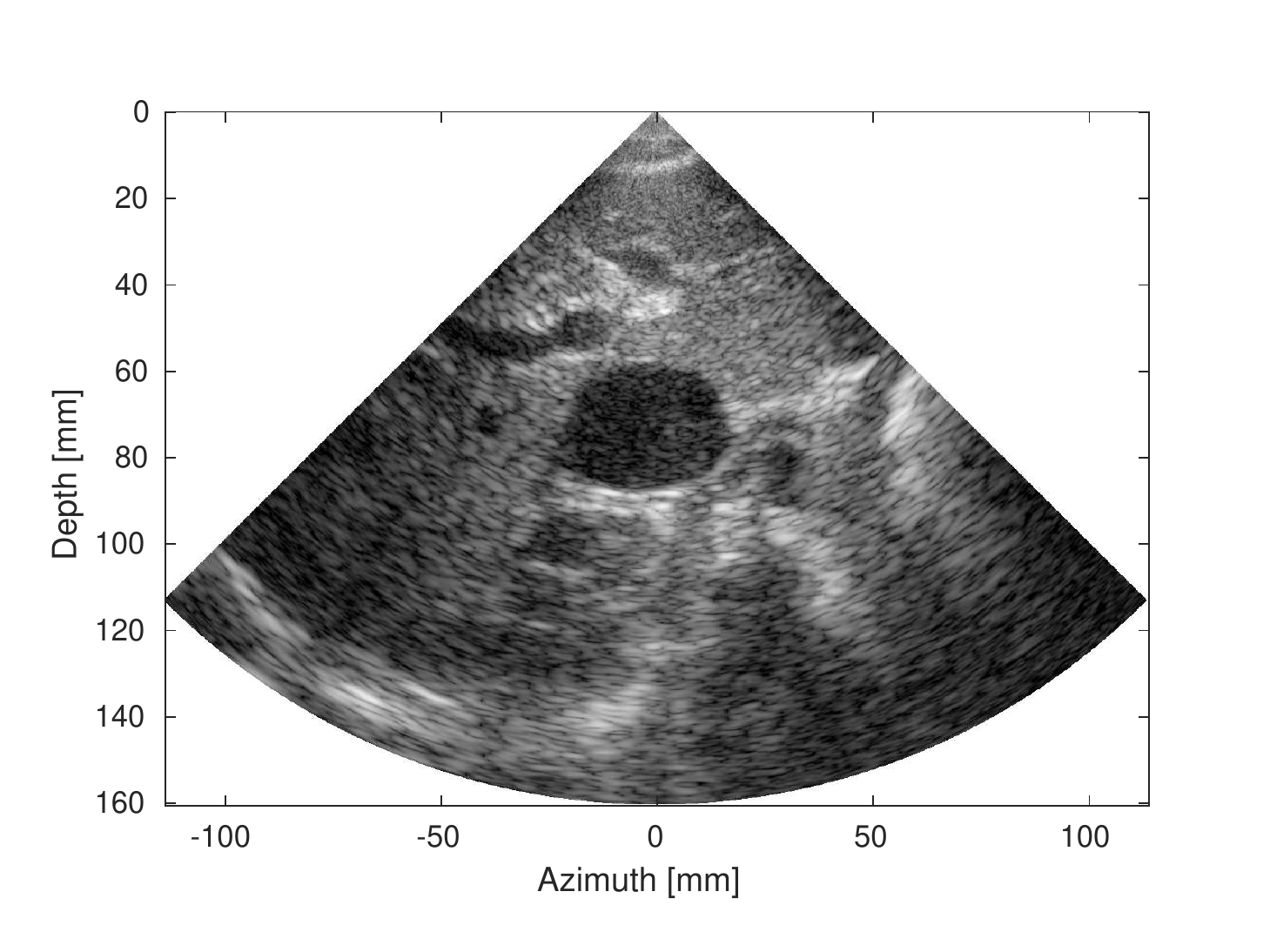} &
		\includegraphics[width = 0.33\textwidth]{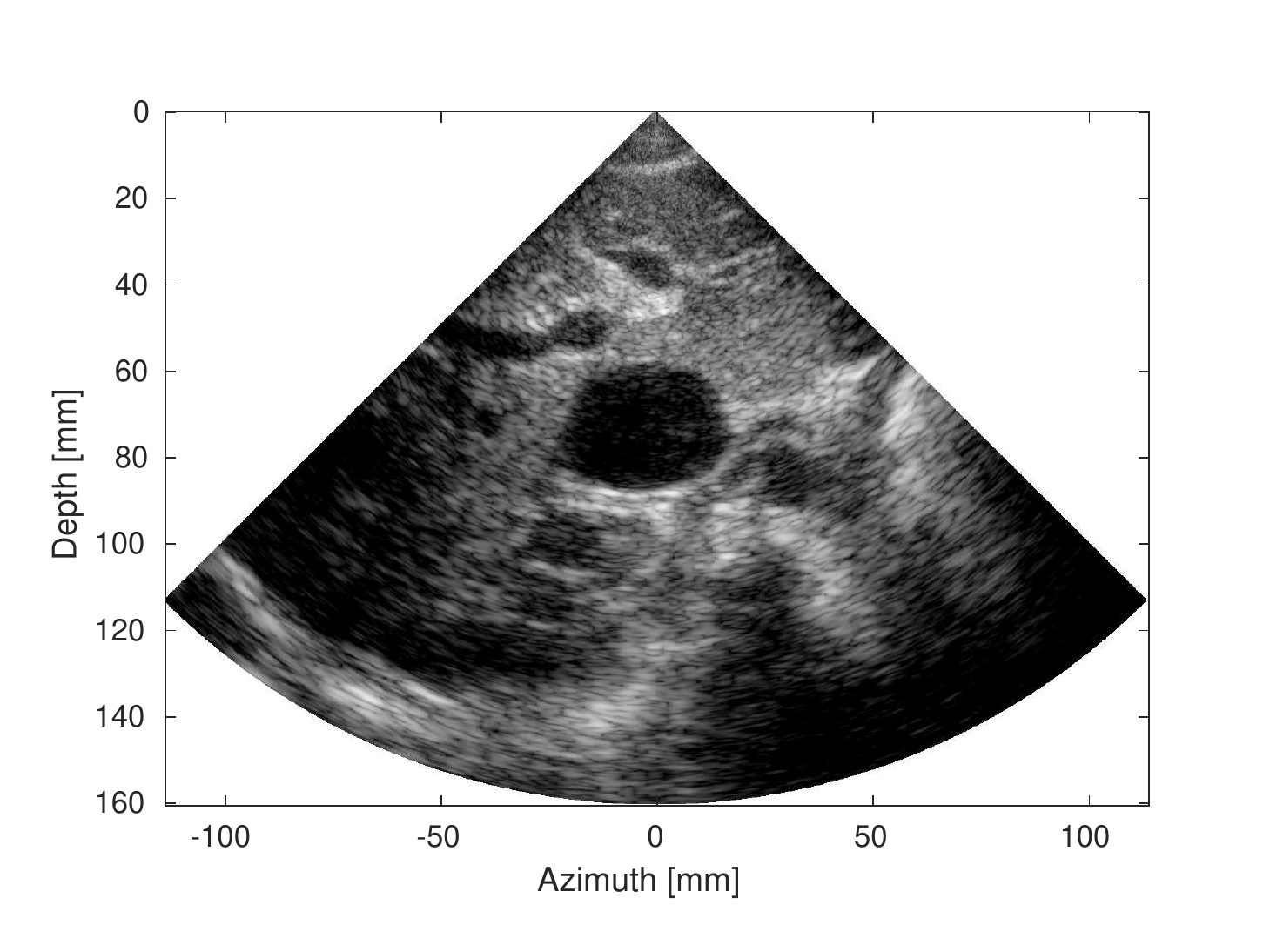}&
		         \\& {\smaller (d)  $6-$MLT, (Tukey, $\alpha$=0.5)} & {\smaller (e) $6-$MLT, CNN}& \\
                & {\smaller CNR=2.44, CR= -23.01dB } & {\smaller CNR= 2.53, CR=-32.81dB } &
                  \\
	\end{tabular}   \\
   \end{minipage}
    \vspace{-0.5cm}
	\caption{\textbf{CNN-based MLT artifact correction tested on {\it{in-vivo}} abdominal frames} (a) an {\it{in-vivo}} frame acquired through SLT from the excluded patient, (b),(d) corresponding $4-$ and $6-$MLT with (Tukey, $\alpha$=0.5) window, and (c),(e) corresponding CNN-corrected frames}
	\label{KidneyFigs}
    \vspace{-0.5cm}
\end{figure}

%

Qualitative evaluation for the phantom frame is presented in S$1$ along with quantitative measurements, provided in the supplementary materials. A magnified region depicts the response from one of the wires of the phantom. A thinner appearance, as compared to the apodized MLT image, can be observed for both $4-$ and $6-$MLT  frames processed with the proposed CNN, since no apodization was needed to attenuate the artifacts. Quantitatively, the CR of the anechoic cyst as compared to the nearby tissue, 
was restored for the case of $6-$MLT, whereas for the $4-$MLT case it was improved by almost $7$ dB as compared to the SLT. Since the network was trained on the data with a higher number of a strong reflectors, thus higher artifact content, it is possible that the artifact content is overestimated in some cases. The images of the bladder (S$2$) appear to have a higher quality in the $4-$MLT and $6-$MLT CNN corrected cases, as compared to the respective apodized versions. Quantitatively, the improvement in contrast over apodized MLT was around $10$ dB for $4$-MLT and $13$ dB for for $6-$MLT.

A slight CNR improvement as compared to the apodized MLT was measured in all cases, except for the $6-$MLT for the tissue mimicking phantom, where the CNR remained the same. The performance of our CNN, verified on the testing set frames of internal organs, and of a tissue mimicking phantom, suggests that it generalizes well to other scenes and patients, despite being trained on a small dataset of bladder frames.  

It should be noted that the coherent processing of the data  (through convolutions applied on the data prior to the envelope detection) along the lateral direction may impose motion artifacts while imaging regions involving rapid movement (such as cardiac tissue and blood).  Nevertheless, in most compensation methods, the correction is performed without relying on the adjacent samples in lateral direction, thus, similar approaches relying on constraints in the lateral direction can be built into the neural network. We defer this case to future studies.

\section{Conclusion}
In this paper, we have demonstrated that correction provided by an end-to-end CNN outperforms the constant apodization-based correction method of MLT cross-talk artifacts, as measured using CR and CNR. 
Moreover, the obtained CNN generalizes well for different anatomical scenes. In the future, we intend to address the problem of MLT artifact suppression for rapidly moving objects scenes, by training a CNN to correct all the lines beamformed from a single transmit event. Furthermore, we aim at exploring the possibility of similarly reconstructing artifact-free images for combined MLT-MLA configurations, that introduce an even larger boost in frame rate. 

\section{Acknowledgments}
This research was partially supported by ERC StG RAPID. 

%
%
%
%
\bibliographystyle{splncs04}
{\footnotesize \bibliography{samplepaper}}

\clearpage

\begin{center}
\textbf{\larger Supplementary material}
\end{center}
 \begin{figure}[h]
\begin{minipage}[]{\linewidth}
	\begin{tabular}{ c@{\hskip 0.001\textwidth}c@{\hskip 0.001\textwidth}c@{\hskip 0.001\textwidth}c} 

		\includegraphics[width = 0.33\textwidth]{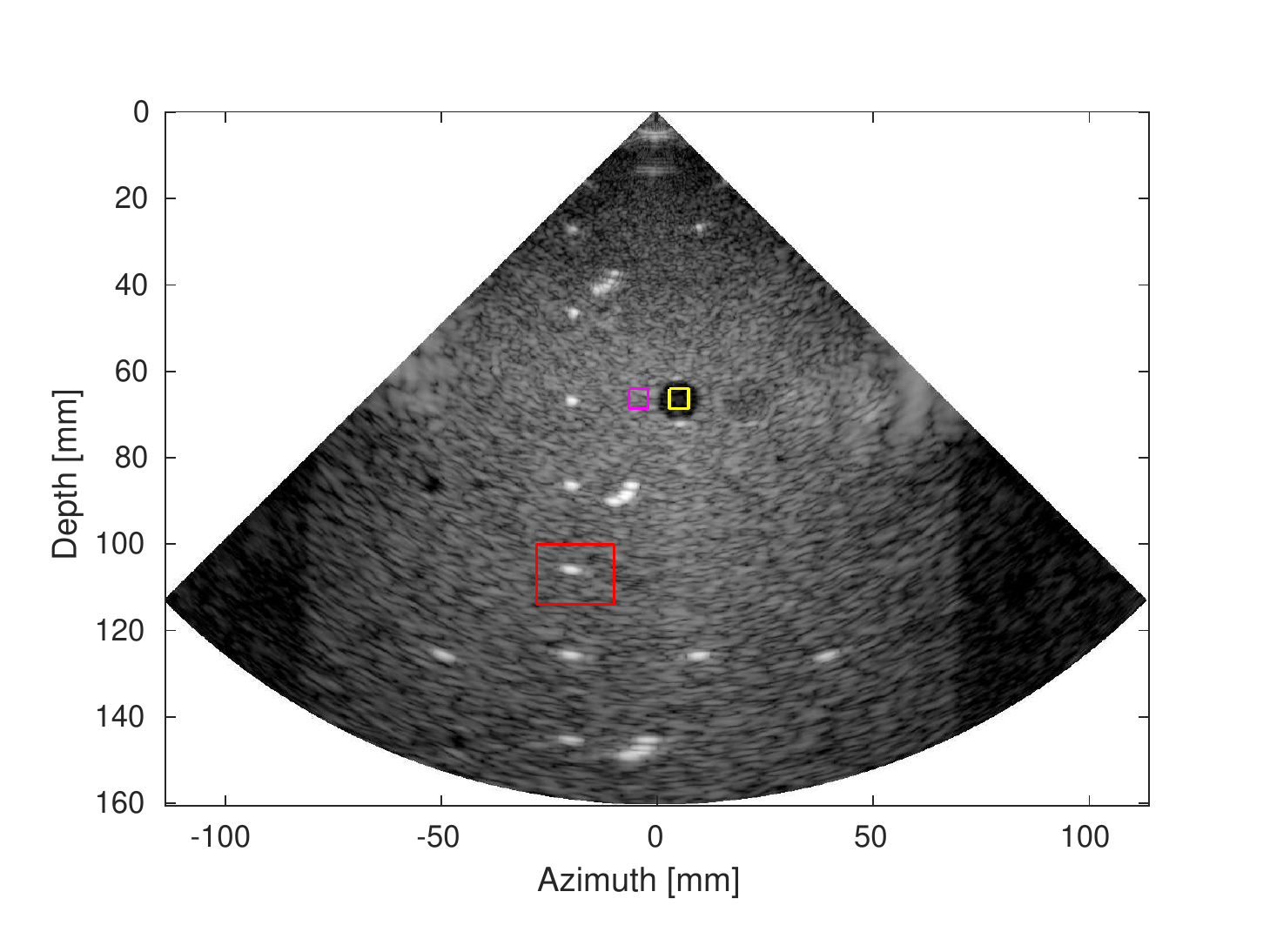} &
		\includegraphics[width = 0.33\textwidth]{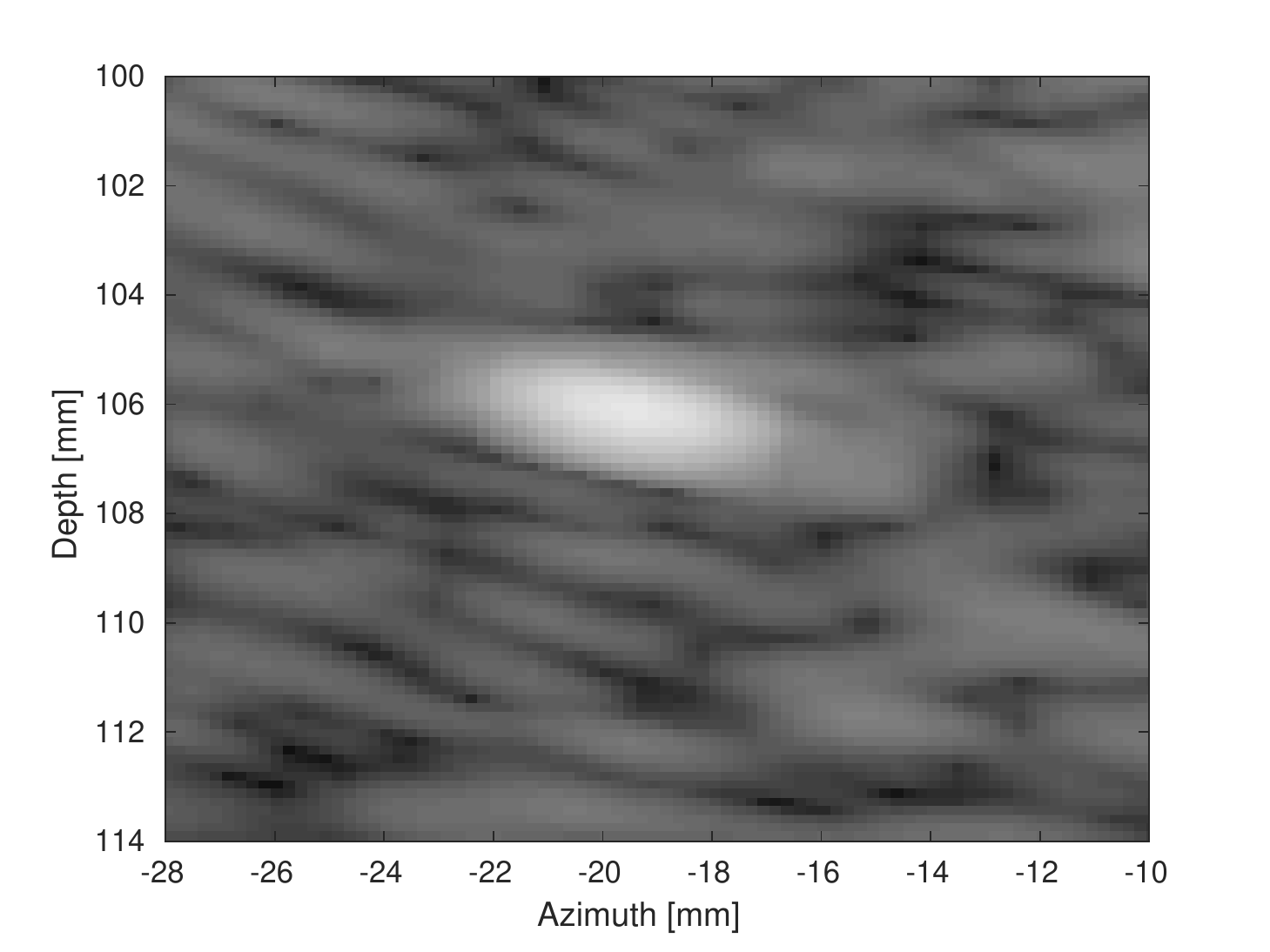} &
		\includegraphics[width = 0.33\textwidth]{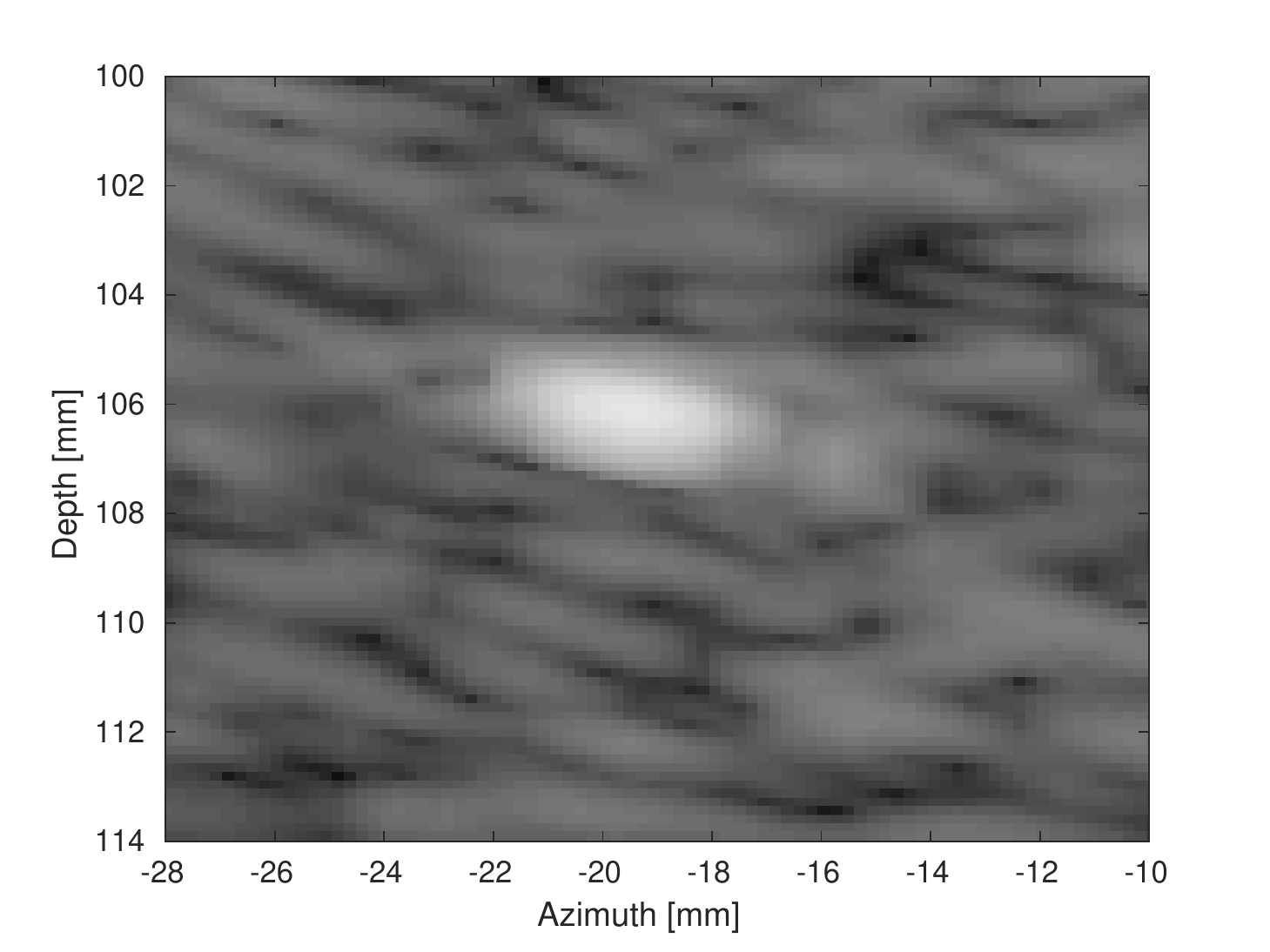} &
	       \\
               (a) SLT & (c) $4-$MLT, (Tukey, $\alpha$=0.5) & (d) $4-$MLT, CNN & \\
        \includegraphics[width = 0.33\textwidth]{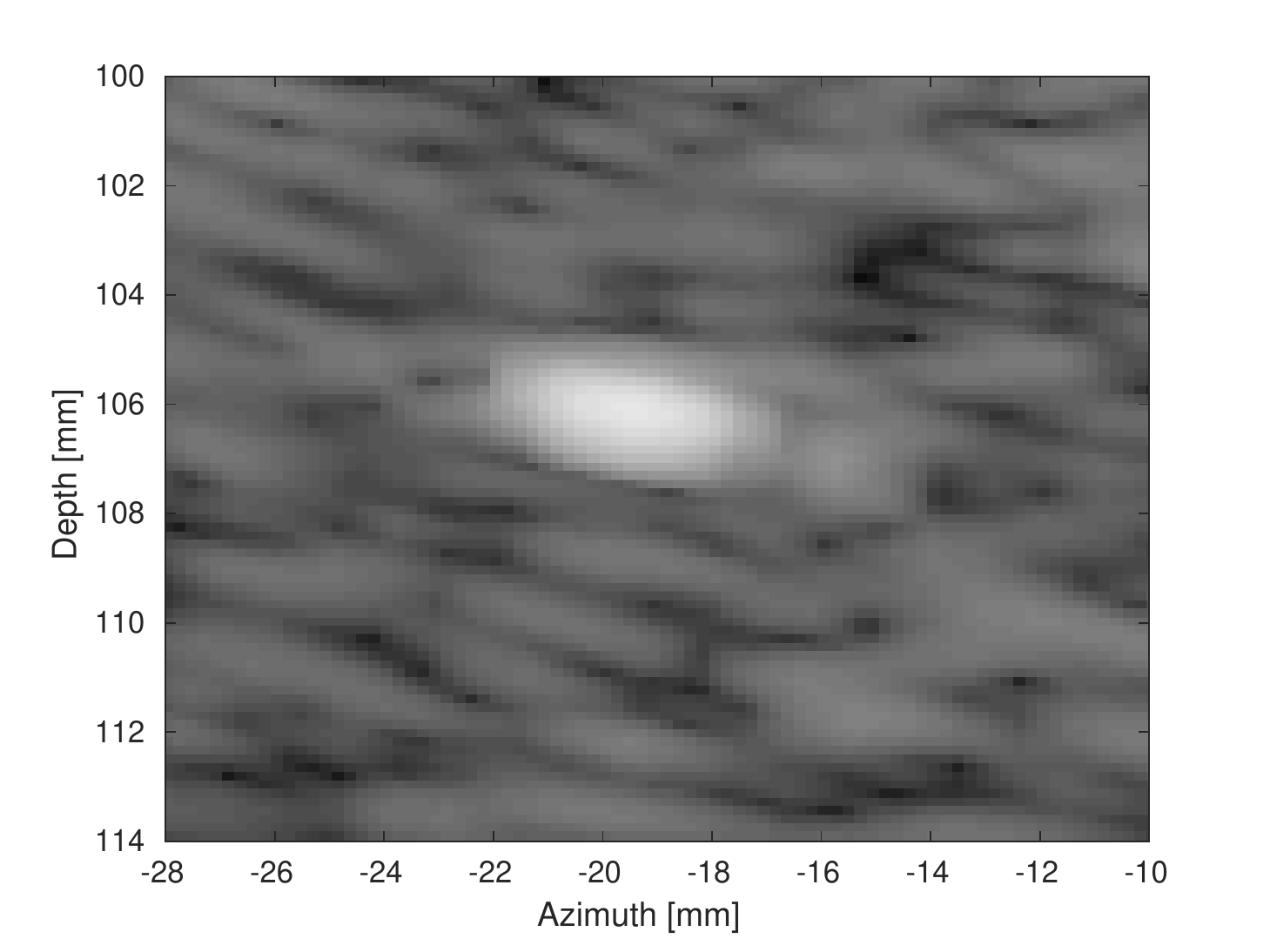}   &
		\includegraphics[width = 0.33\textwidth]{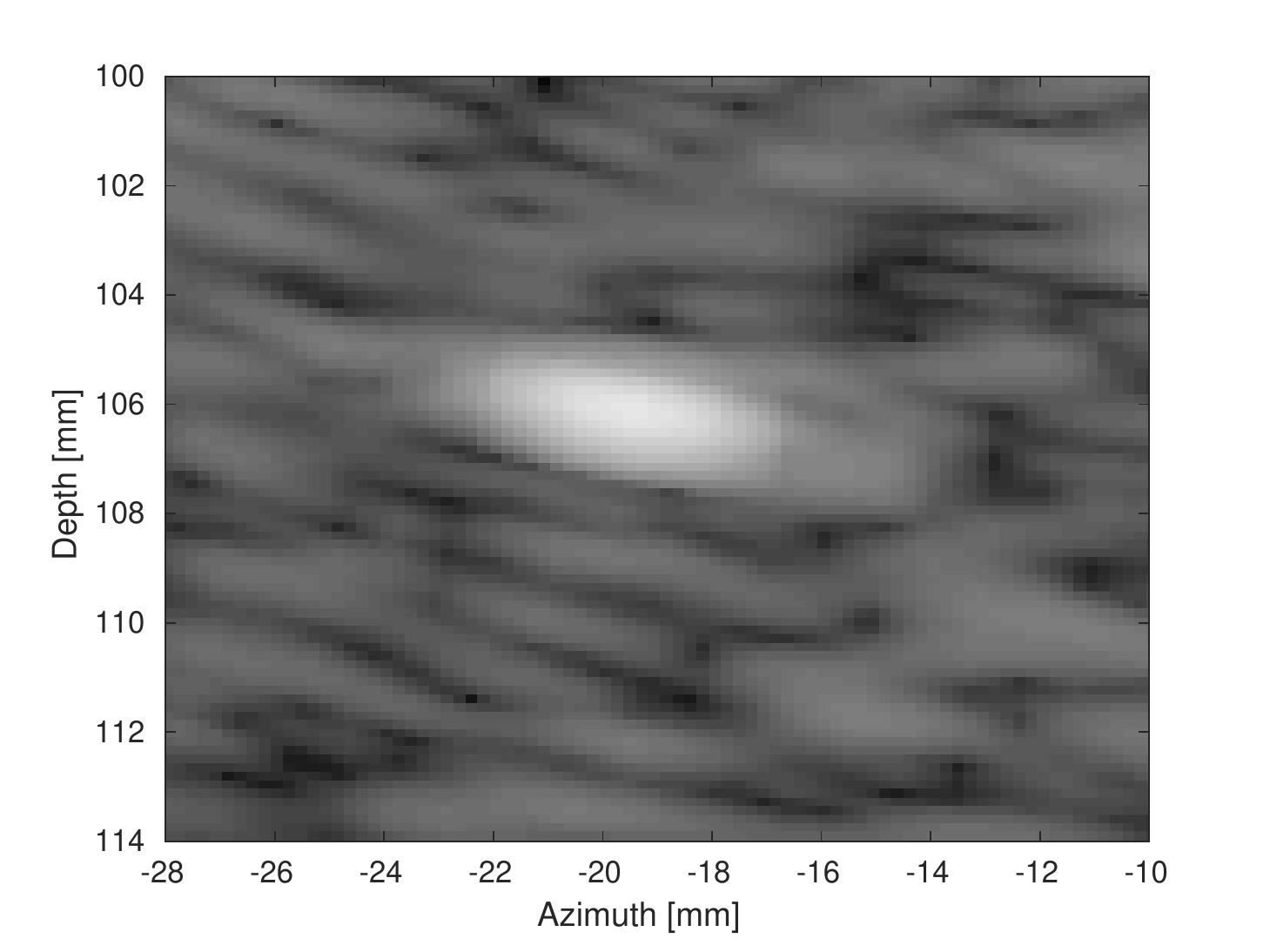} &
		\includegraphics[width = 0.33\textwidth]{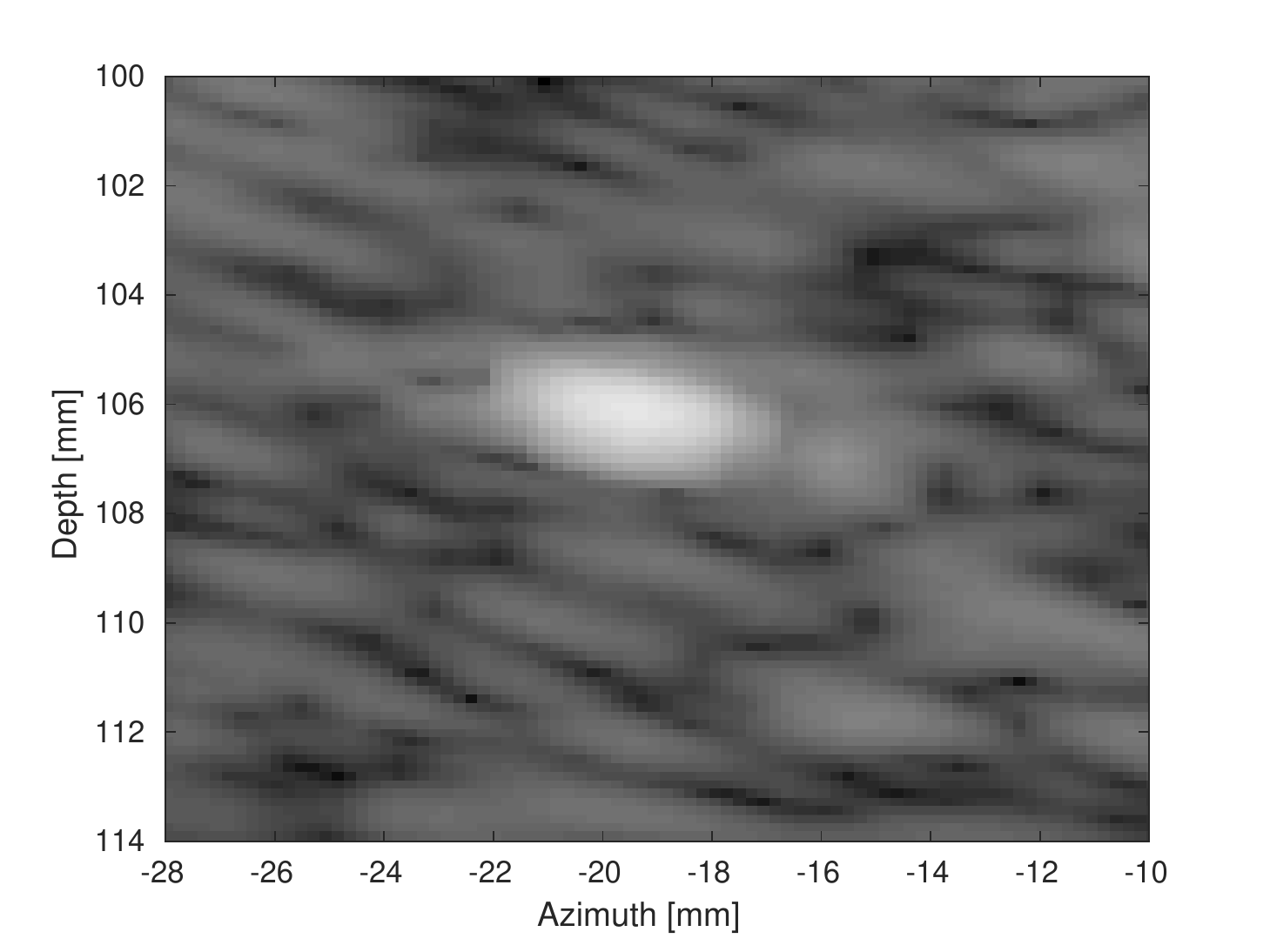}&
		         \\
      (b) SLT zoom-in & (e) $6-$MLT, (Tukey, $\alpha$=0.5) & (f) $6-$MLT, CNN& \\
	\end{tabular}   \\
    \begin{center}

    \resizebox{.8\textwidth}{!}{\begin{tabular}{|c |c |c |c |c| c |c|} 
 \hline
  &\multicolumn{1}{|c|}{{\smaller SLT}}&\multicolumn{2}{|c|}{\smaller $4-$MLT}&\multicolumn{2}{|c|}{\smaller $6-$MLT}\\ 
\hline
 & {\smaller }  & {\smaller (Tukey, $\alpha$=0.5)} & {\smaller CNN} & {\smaller (Tukey, $\alpha$=0.5)} & {\smaller CNN} \\ [0.5ex] 
 \hline
{\smaller CNR}  & {\smaller $1.67$}  & {\smaller $1.65$} & {\smaller $1.71$}  & {\smaller $1.64$} & {\smaller $1.64$}  \\ 
\hline
{\smaller CR(dB)}  & {\smaller $-24.27$}   & {\smaller $-22.78$} & {\smaller $-31.18$} & {\smaller $-22.68$} & {\smaller $-24.68$} \\ 
\hline
\end{tabular}
}
\vspace{1mm}
\label{Table1}
\end{center}
\vspace{-1cm}
   \end{minipage}
   
    \vspace{0.5cm}
	\caption*{\small \textbf{S1. CNN-based MLT artifact correction tested on phantom data.} (a) a phantom frame acquired through SLT. (b) the zoomed-in region marked in red box in (a). (c)-(f) corresponding zoomed-in regions from MLT with (Tukey, $\alpha$=0.5)/CNN-Corrected frames. The table summarizes the CNR and CR measures for each frame, calculated with respect to the marked (in a) cyst region (yellow) and background region (magenta). }
	\label{PhantomFigs}
\end{figure}


\begin{figure}[h]
\begin{minipage}[]{\linewidth}
	\begin{tabular}{ c@{\hskip 0.001\textwidth}c@{\hskip 0.001\textwidth}c@{\hskip 0.001\textwidth}c} 

		\includegraphics[width = 0.33\textwidth]{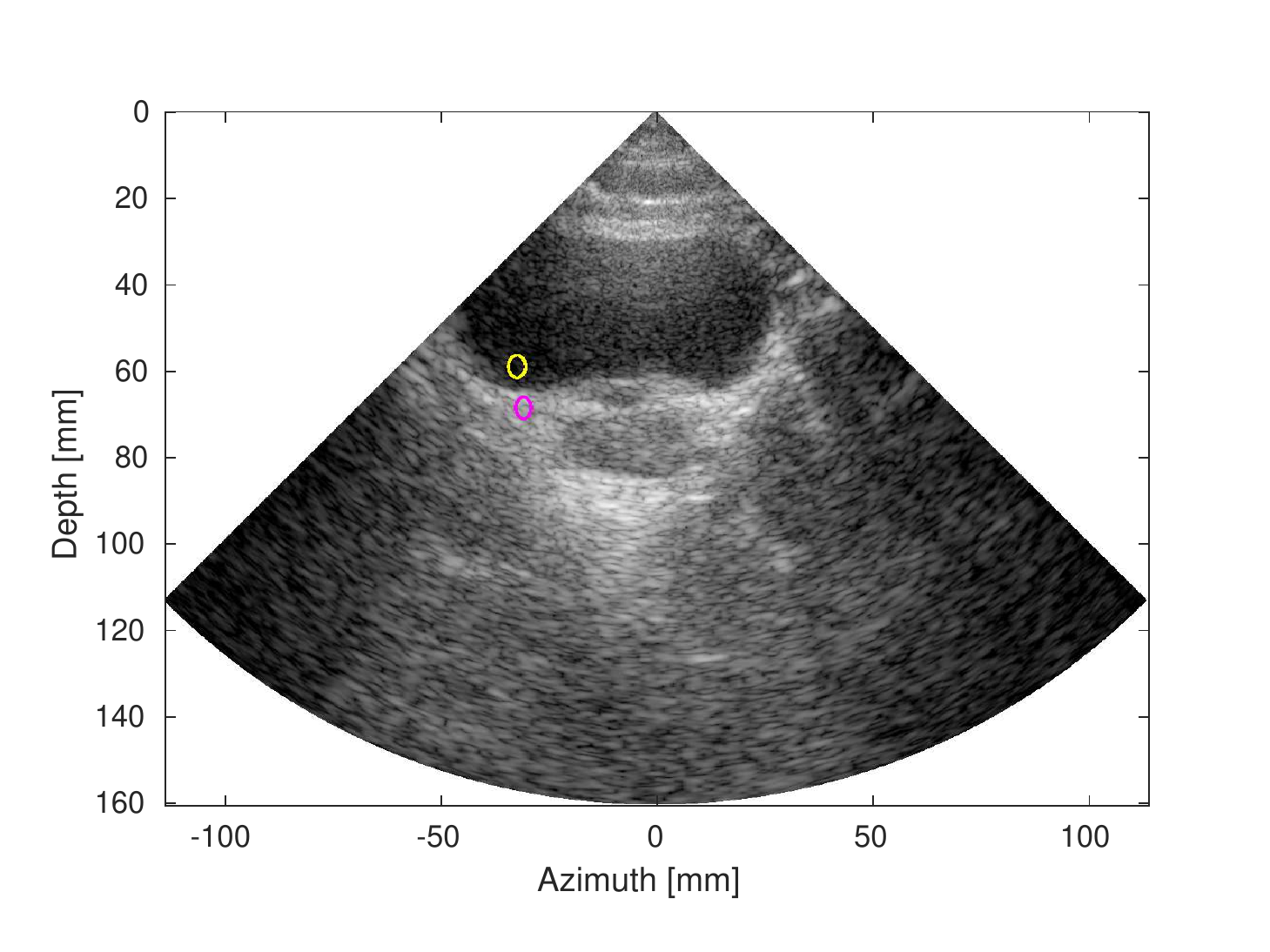} &
		\includegraphics[width = 0.33\textwidth]{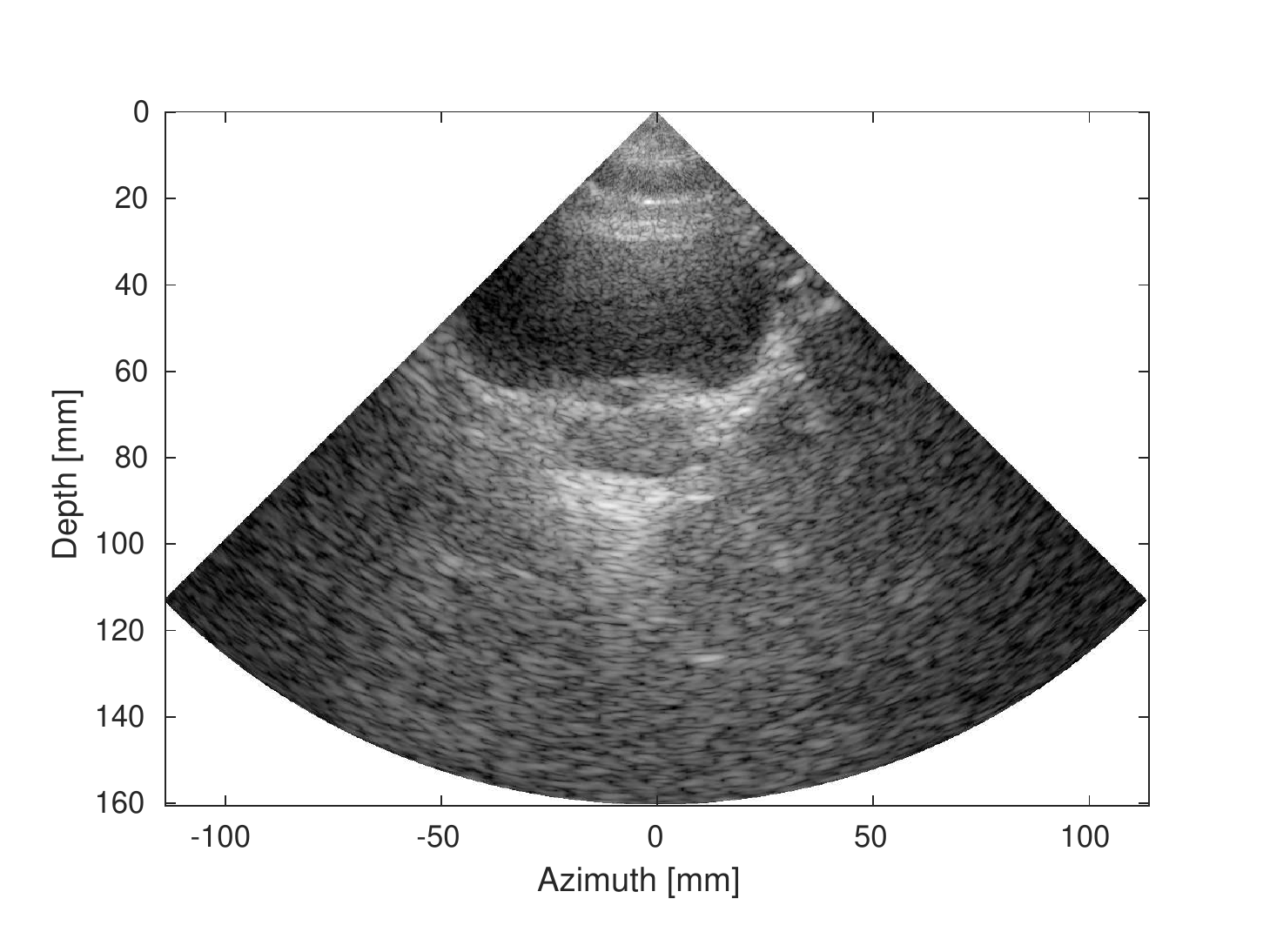} &
		\includegraphics[width = 0.33\textwidth]{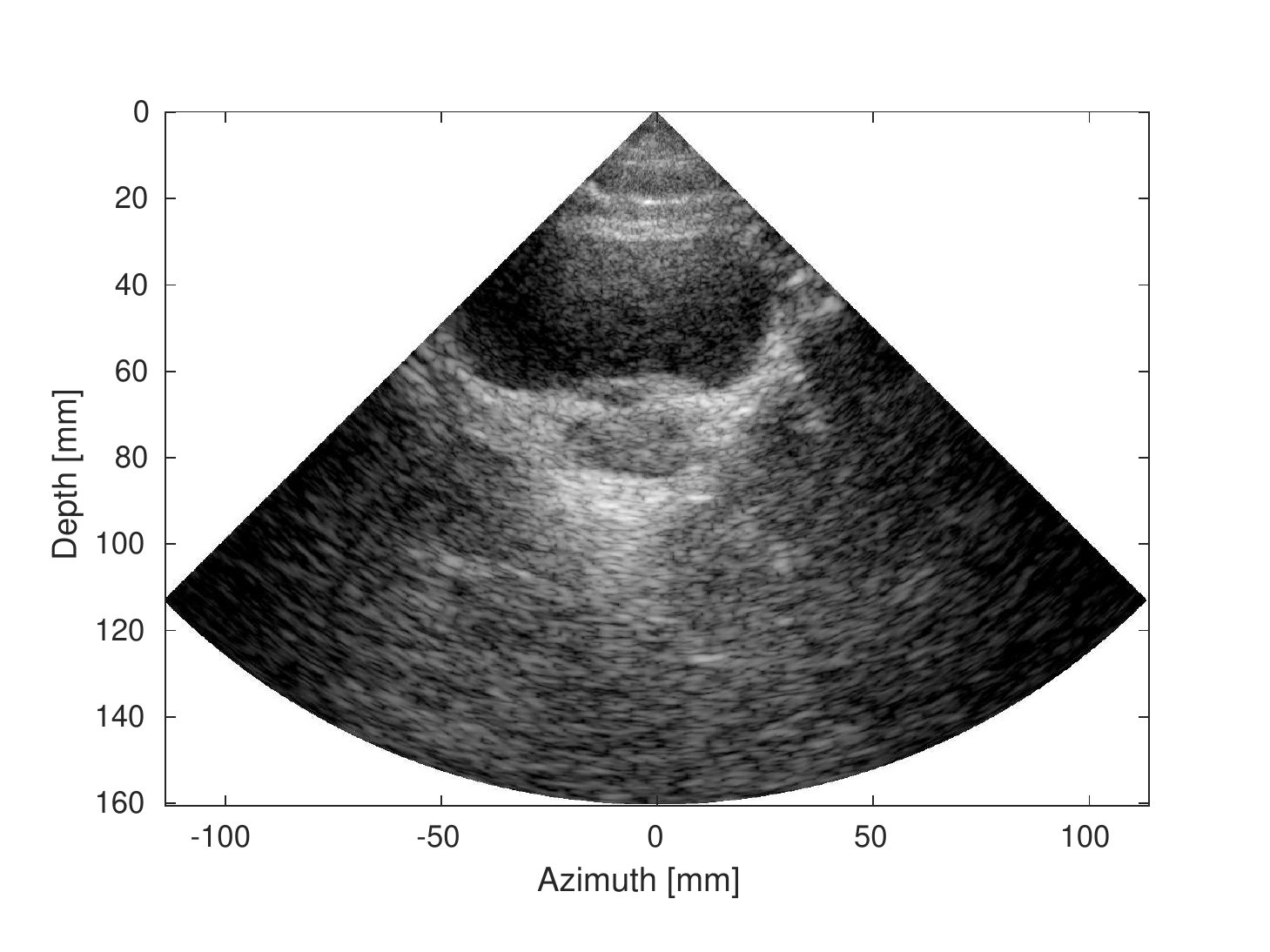} &
	       \\
              {\smaller (a) SLT} & {\smaller (b) $4-$MLT, (Tukey, $\alpha$=0.5)} & (c) {\smaller  $4-$MLT, CNN}  \\
               {\smaller CNR=2.33, CR=-29.33dB} & {\smaller CNR=1.69, CR=-19.16dB } & {\smaller CNR=2.2, CR= -29.24dB} &\\
          \includegraphics[width = 0.33\textwidth]{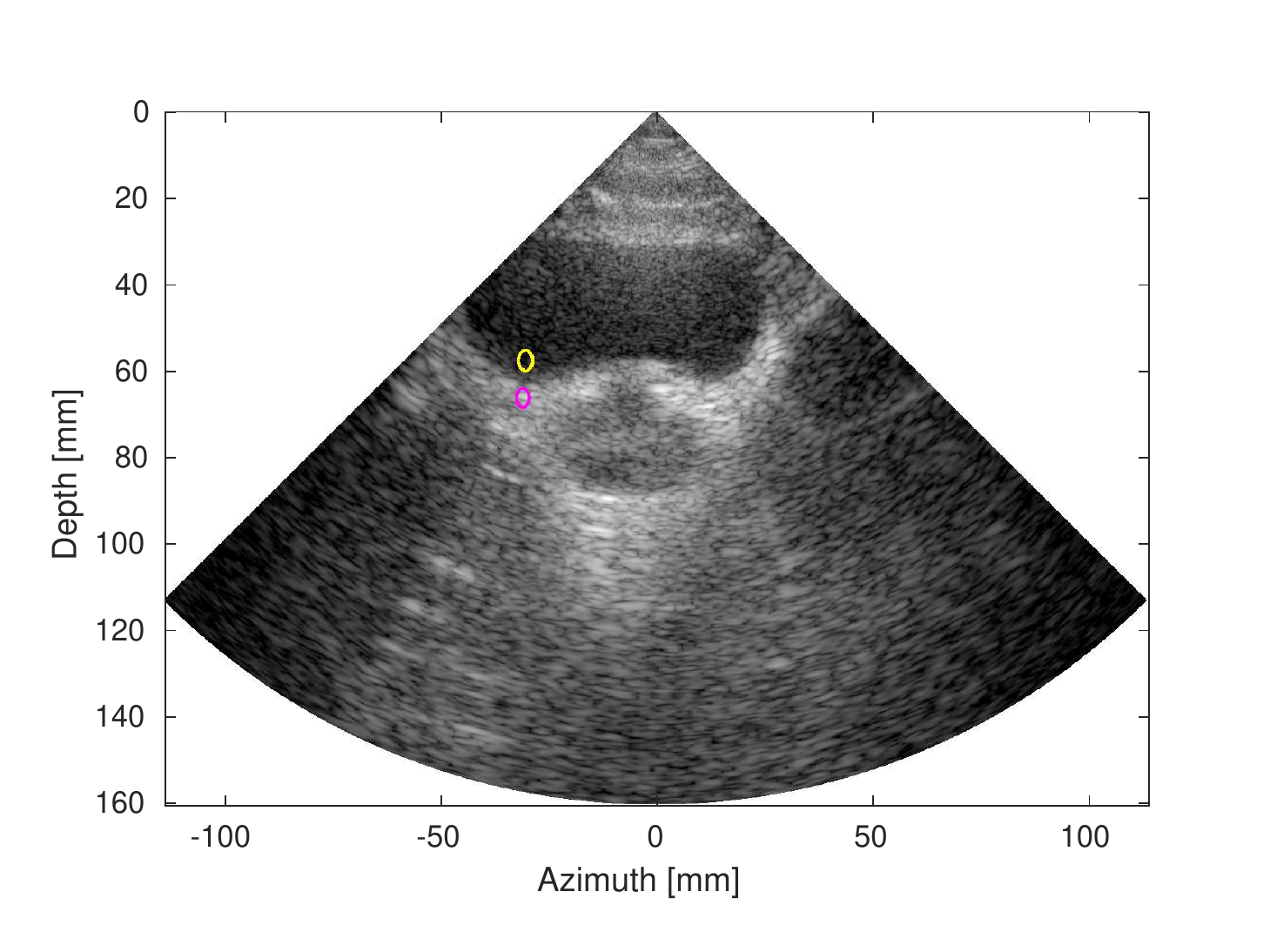} &
		\includegraphics[width = 0.33\textwidth]{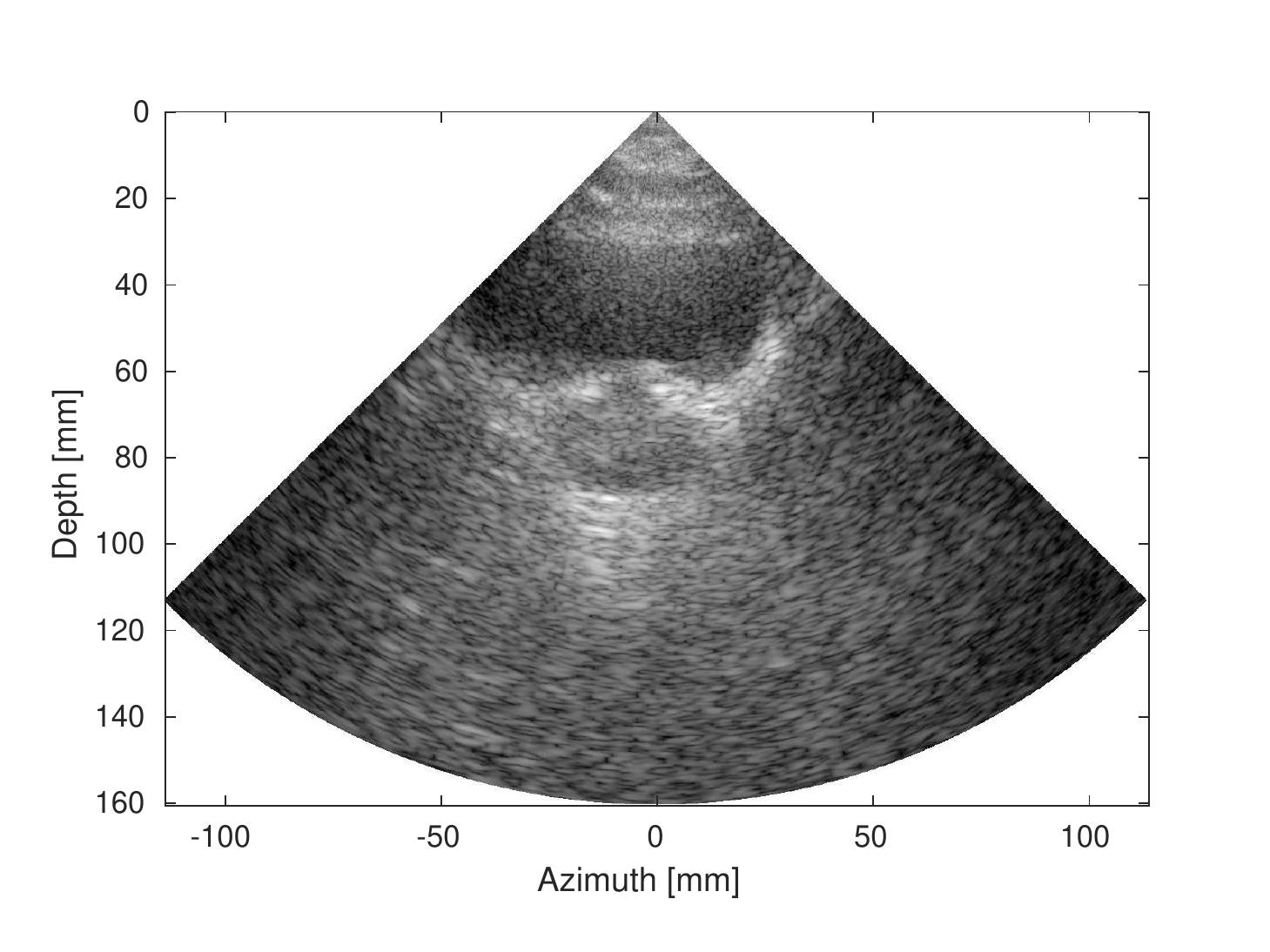} &
		\includegraphics[width = 0.33\textwidth]{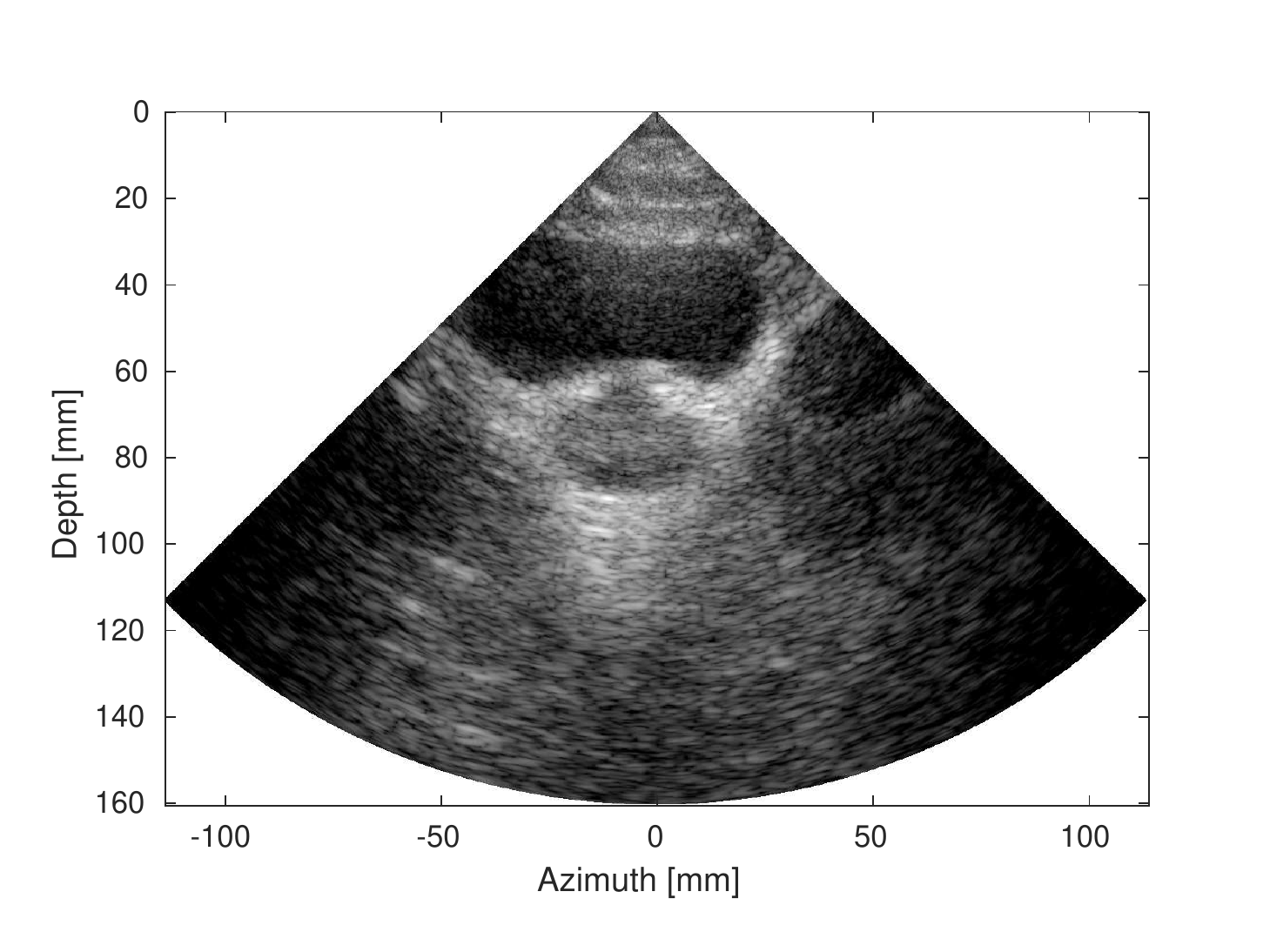}&
		         \\
                {\smaller (d) SLT} &{\smaller (e)  $6-$MLT, (Tukey, $\alpha$=0.5)} & {\smaller (f)  $6-$MLT, CNN}& \\
                {\smaller CNR=1.41, CR=-27.37dB } & {\smaller CNR=1.19, CR=-12.99dB } & {\smaller CNR=1.41, CR=-25.96dB } &\\
	\end{tabular}   \\
   \end{minipage}
    \vspace{-0.3cm}
	\caption*{\small \textbf{S2. CNN-based MLT artifact correction tested on {\it{in-vivo}} bladder frames} (a),(d) an {\it{in-vivo}} frames acquired through SLT (b),(e) corresponding 4- and 6-MLT with (Tukey, $\alpha$=0.5) window, and (c),(f) corresponding CNN-corrected frames}
	\label{BladderFigs}
\end{figure}

\end{document}